\documentclass{article} 

\usepackage{enumitem}

\usepackage[final]{neurips_2025}

\usepackage{amsmath,amssymb,amsthm}
\usepackage{bm}
\usepackage{natbib}
\usepackage{graphicx,here,comment}
\usepackage{algorithmic}
\usepackage{algorithm}
\usepackage{xcolor}
\usepackage{wrapfig}
\usepackage{footnote}
\usepackage{multirow}
\makesavenoteenv{table}  

\usepackage{mathtools}

\usepackage{hyperref}

\newtheorem{theorem}{Theorem}

\newtheorem{lemma}[theorem]{Lemma}

\newtheorem{assumption}{Assumption}
\theoremstyle{definition}

\newcommand{\rbr}[1]{\left(#1\right)}
\newcommand{\sbr}[1]{\left[#1\right]}
\newcommand{\cbr}[1]{\left\{#1\right\}}

\newcommand{\R}{\mathbb{R}}
\newcommand{\N}{\mathbb{N}}

\newcommand{\mH}{\mathcal{H}}

\newcommand{\mT}{\mathcal{T}}

\newcommand{\mX}{\mathcal{X}}

\newcommand{\mE}{\mathcal{E}}

\renewcommand{\hat}{\widehat}
\renewcommand{\tilde}{\widetilde}

\newcommand{\argmax}{\operatornamewithlimits{argmax}}

\def\bX{\mathbf{X}}

\def\bK{\mathbf{K}}

\def\bk{\bm{k}}

\def\bI{\bm{I}}

\def\bx{\bm{x}}

\usepackage{color}

\usepackage{newtxtext,newtxmath}

\newcommand{\secref}[1]{Section~\ref{#1}}
\newcommand{\appref}[1]{Appendix~\ref{#1}}
\newcommand{\algoref}[1]{Algorithm~\ref{#1}}
\newcommand{\asmpref}[1]{Assumption~\ref{#1}}
\newcommand{\thmref}[1]{Theorem~\ref{#1}}
\newcommand{\lemref}[1]{Lemma~\ref{#1}}

\newcommand{\figref}[1]{Figure~\ref{#1}}

\newcommand{\1}{\mbox{1}\hspace{-0.25em}\mbox{l}}

\newcommand{\sek}{k_{\mathrm{SE}}}
\newcommand{\matk}{k_{\mathrm{Mat\acute{e}rn}}}
\newcommand{\cse}{C_{\mathrm{SE}}}
\newcommand{\cmat}{C_{\mathrm{Mat}}}
\newcommand{\cset}{\tilde{C}_{\mathrm{SE}}}
\newcommand{\cmatt}{\tilde{C}_{\mathrm{Mat}}}
\newcommand{\tse}{\overline{T}_{\mathrm{SE}}}
\newcommand{\tmat}{\overline{T}_{\mathrm{Mat}}}
\newcommand{\utse}{\underline{T}_{\mathrm{SE}}}
\newcommand{\utmat}{\underline{T}_{\mathrm{Mat}}}
\newcommand{\utlse}{\underline{T}_{\mathrm{SE}}^{(\lambda)}}
\newcommand{\utlmat}{\underline{T}_{\mathrm{Mat}}^{(\lambda)}}
\newcommand{\ulse}{\underline{\lambda}_{\mathrm{SE}}}
\newcommand{\ulmat}{\underline{\lambda}_{\mathrm{Mat}}}

\title{Gaussian Process Upper Confidence Bound Achieves Nearly Optimal Regret \\ in Noise-Free Gaussian Process Bandits}

\author{%
  Shogo Iwazaki \\
  LY Corporation \\
  Tokyo, Japan \\
  \texttt{siwazaki@lycorp.co.jp} \\
}

\begin{document}
\maketitle

\begin{abstract}
We study the noise-free Gaussian Process (GP) bandit problem, in which a learner seeks to minimize regret through noise-free observations of a black-box objective function that lies in a known reproducing kernel Hilbert space (RKHS).
The Gaussian Process Upper Confidence Bound (GP-UCB) algorithm is a well-known approach for GP bandits, where query points are adaptively selected based on the GP-based upper confidence bound score.
While several existing works have reported the practical success of GP-UCB, its existing regret guarantees remain suboptimal. However, GP-UCB often empirically outperforms other nearly-optimal noise-free algorithms that use non-adaptive sampling schemes.
This paper resolves the gap between theoretical and empirical performance by establishing a nearly-optimal regret upper bound for noise-free GP-UCB. Specifically, our analysis provides the first constant cumulative regret bounds in the noise-free setting for both the squared exponential kernel and the Mat\'ern kernel with some degree of smoothness.
\end{abstract}

\section{Introduction}
\begin{figure}
    \centering
    \includegraphics[width=0.32\linewidth]{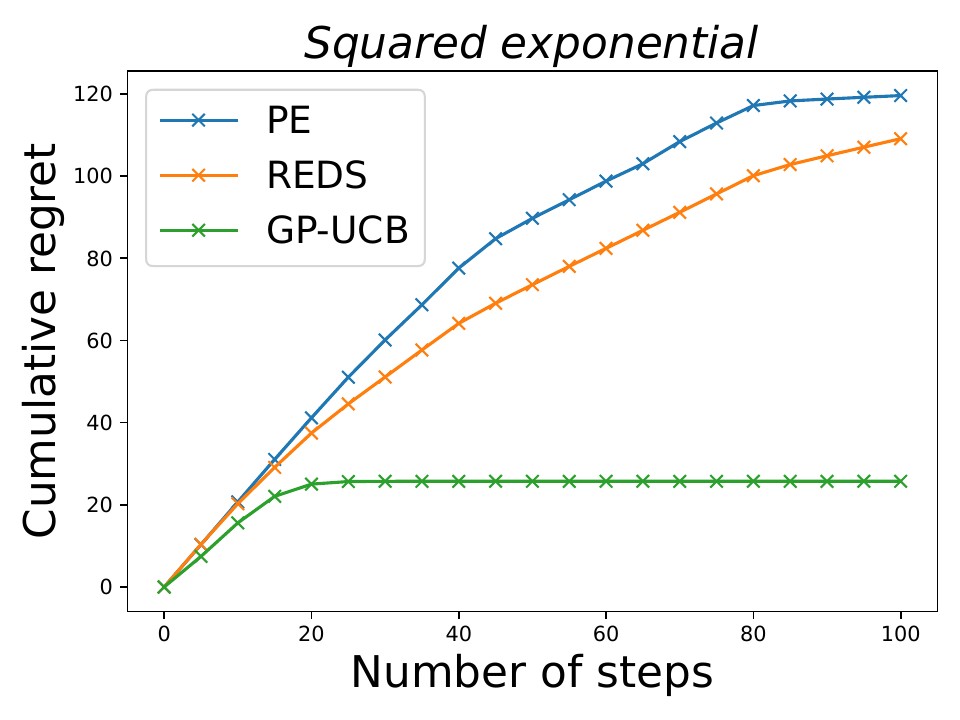}
    \includegraphics[width=0.32\linewidth]{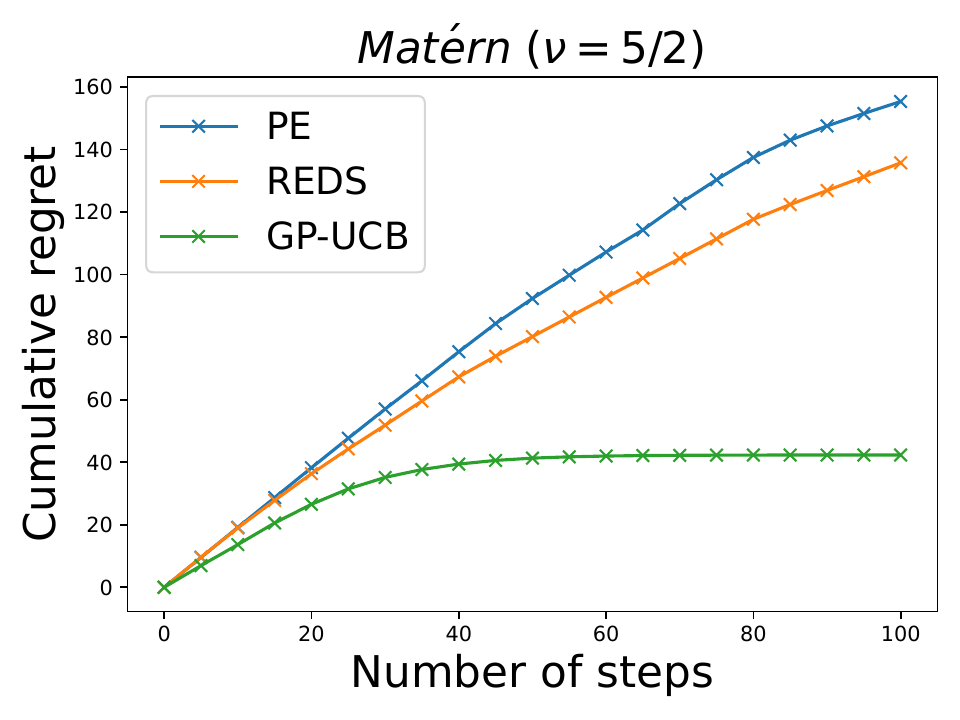}
    \includegraphics[width=0.32\linewidth]{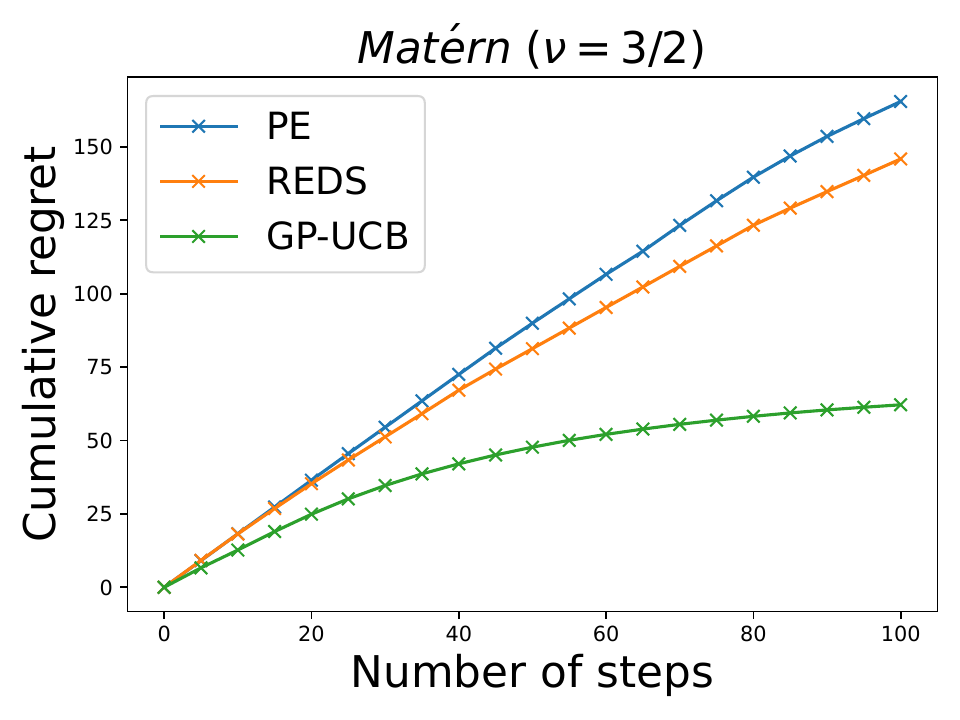}
    \caption{Empirical performance comparison of GP-UCB and two existing nearly-optimal algorithms: random exploration with domain shrinking (REDS)~\citep{salgiarandom} and phased elimination (PE)~\citep{iwazaki2025improvedregretanalysisgaussian}. 
    From left to right, the plots show the average cumulative regret over $3000$ independent runs under the squared exponential kernel, Mat\'ern kernel ($\nu = 5/2$), and Mat\'ern kernel ($\nu = 3/2$) with $d = 2$, respectively. Here, $d$ and $\nu$ represent the dimension of the input and the smoothness parameter, respectively.
    Detailed experimental settings are provided in \appref{sec:exp_detail}.
    }
    \label{fig:emp_regret}
\end{figure}

This paper studies the noise-free Gaussian Process (GP) bandit problem, where the learner seeks to minimize regret through noise-free observations of the black-box objective function. Several existing works tackle this problem, and some of them~\citep{iwazaki2025improvedregretanalysisgaussian,salgiarandom} propose algorithms whose regret nearly matches the lower bound of \citep{liregret}. For ease of theoretical analysis, these algorithms rely on the non-adaptive sampling scheme, whose query points are chosen independently of the observed function values, such as the uniform sampling~\citep{salgiarandom} or maximum variance reduction~\citep{iwazaki2025improvedregretanalysisgaussian}. 
Although the theoretical superiority of such non-adaptive algorithms is shown, in existing noisy setting literature~\citep{bogunovic2022robust,iwazakinear,li2022gaussian}, their empirical performance has been reported to be worse than that of fully adaptive strategies such as GP upper confidence bound (GP-UCB)~\citep{srinivas10gaussian}.
Unsurprisingly, we also observe such empirical and practical gaps in a noise-free setting, as shown in \figref{fig:emp_regret}.
These observations suggest the possibility of further theoretical improvement in the practical fully adaptive algorithm.
From this motivation, our work aims to establish the nearly-optimal regret for GP-UCB, which is one of the well-known adaptive GP bandit algorithms, and its existing guarantees only show strictly sub-optimal regret in a noise-free setting~\citep{kim2024bayesian,lyu2019efficient}. 

\paragraph{Contributions.} Our contributions are summarized below:
\begin{itemize}
    \item We give a refined regret analysis of GP-UCB (Theorems~\ref{thm:cr_gp-ucb} and \ref{thm:sr_gp-ucb}), which matches both the cumulative regret lower bounds of \citep{liregret} and the simple regret lower bound of \citep{bull2011convergence} up to polylogarithmic factors in the Mat\'ern kernel. Regarding cumulative regret, our analysis shows that GP-UCB achieves constant $O(1)$ regret under the squared exponential and Mat\'ern kernel with $d < \nu$. The results are summarized in Tables~\ref{tab:nl_cr_compare} and \ref{tab:nl_sr_compare}.
    \item Our key theoretical contribution consists of new algorithm-independent upper bounds for the observed posterior standard deviations (Lemmas~\ref{lem:pv_ub}--\ref{lem:cpv_ub}) by bridging the information gain-based analysis in the noisy regime to the noise-free setting.
    Furthermore, as discussed in \secref{sec:reg_ub_gpucb}, these results have the potential to translate existing confidence bound-based algorithms for noisy settings into nearly-optimal noise-free variants beyond the analysis of GP-UCB.
\end{itemize}

\paragraph{Related works.} 
Various existing works study the theory of noisy GP bandits~\citep{chowdhury2017kernelized,li2022gaussian,scarlett2017lower,srinivas10gaussian,valko2013finite}. 
Regarding noise-free settings, to our knowledge, \citep{bull2011convergence} is the first work that shows both the upper bound and the lower bound for simple regret via the expected improvement (EI) strategy. After that, cumulative regret bounds are established in \citep{lyu2019efficient} with GP-UCB. 
Recently, \citet{vakili2022open} conjectures the lower bound of the cumulative regret in the noise-free setting, suggesting a superior algorithm exists in the noise-free setting.
Later, the following work~\citep{liregret} formally validates the conjectured lower bound of \citep{vakili2022open}~\footnote{Although \citet{liregret} considers the cascading structure of the observation process, the standard noise-free setting is a special case of their setting.}.
Motivated by the lower bounds, several works~\citep{flynn2024tighter,kim2024bayesian,iwazaki2025improvedregretanalysisgaussian,salgiarandom} studied improved algorithms to achieve superior regret to the result of \citep{lyu2019efficient}.
Although some of them propose the nearly-optimal algorithms~\citep{iwazaki2025improvedregretanalysisgaussian,salgiarandom}, their algorithms are based on a non-adaptive sampling scheme, whose inferior performance has been reported in the existing work~\citep{bogunovic2022robust,iwazakinear,li2022gaussian}. 
Here, our motivation is to establish the nearly-optimal theory under the fully adaptive nature of GP-UCB; however, our proof relates to the analysis in \citep{iwazaki2025improvedregretanalysisgaussian} for the non-adaptive maximum variance reduction algorithm. Their analysis includes the noise-free setting as a special case and provides nearly-optimal regret under broader varying noise variance settings. However, its applicability is limited to the maximum variance reduction algorithm. Our analysis can be interpreted as a refined version of that in \citep{iwazaki2025improvedregretanalysisgaussian} by focusing on the noise-free setting. 
Finally, although our paper studies the frequentist assumption that the underlying function is fixed, our core results (Lemmas~\ref{lem:pv_ub}--\ref{lem:cpv_ub}) are also applicable to the regret analysis in the Bayesian setting, whose underlying function is drawn from a known GP~\citep{freitas2012exponential,grunewalder2010regret,Russo2014-learning,russo2014learning,scarlett2018tight,srinivas10gaussian}.

\begin{table}[tb]
    \centering
    \caption{Comparison between existing noise-free algorithms' guarantees for cumulative regret and our result (adapted from \citep{iwazaki2025improvedregretanalysisgaussian}). 
    As with the table in \citep{iwazaki2025improvedregretanalysisgaussian}, the smoothness parameter $\nu$ of the Mat\'ern kernel and $\alpha > 0$ in PE are assumed to be $\nu > 1/2$ and an arbitrary fixed constant, respectively. 
    Furthermore, all the parameters ($d$, $\ell$, $\nu$, and $B$) except for $T$ are assumed to be $\Theta(1)$.
    The ``Type'' column indicates whether the regret guarantee is (D)eterministic or (P)robabilistic. 
    Here, a regret bound is labeled “deterministic” if it holds for all possible realizations of the inputs, without relying on probabilistic assumptions.
    Here, $\tilde{O}(\cdot)$ hides polylogarithmic factors in $T$.
    }
    \begin{tabular}{c|c|c|c|c|c}
    \multicolumn{1}{c|}{\multirow{2}{*}{Algorithm}} & \multicolumn{1}{|c|}{\multirow{2}{*}{Regret (SE)}} & \multicolumn{3}{|c|}{Regret (Mat\'ern)} & \multirow{2}{*}{Type} \\ \cline{3-5}
    \multicolumn{1}{c|}{}  & \multicolumn{1}{|c|}{}  & $\nu < d$  & $\nu = d$  & $\nu > d$ &  \\ \hline \hline
     GP-UCB & \multirow{3}{*}{$O\rbr{\sqrt{T \ln^{d} T}}$} & \multicolumn{3}{|c|}{\multirow{3}{*}{$\tilde{O}\rbr{T^{\frac{\nu + d}{2\nu + d}}}$}} & \multirow{3}{*}{D} \\ 
     \citep{lyu2019efficient} & & \multicolumn{3}{|c|}{} & \\ 
     \citep{kim2024bayesian} & & \multicolumn{3}{|c|}{} & \\ \hline
     Explore-then-Commit & \multirow{2}{*}{N/A} & \multicolumn{3}{|c|}{\multirow{2}{*}{$\tilde{O}\rbr{T^{\frac{d}{\nu + d}}}$}} & \multirow{2}{*}{P} \\ 
     \citep{vakili2022open} &  & \multicolumn{3}{|c|}{} & \\ \hline
         Kernel-AMM-UCB
      & \multirow{2}{*}{$O\rbr{\ln^{d+1} T}$} & \multicolumn{3}{|c|}{\multirow{2}{*}{$\tilde{O}\Bigl(T^{\frac{\nu d + d^2}{2\nu^2 + 2\nu d + d^2}}\Bigr)$}} & \multirow{2}{*}{D} \\ 
      \citep{flynn2024tighter}
      &  & \multicolumn{3}{|c|}{} & \\ \hline
     REDS & \multirow{2}{*}{N/A} & \multirow{2}{*}{$\tilde{O}\rbr{T^{\frac{d - \nu}{d}}}$} & \multirow{2}{*}{$O\rbr{\ln^{\frac{5}{2}} T}$} & \multirow{2}{*}{$O\rbr{\ln^{\frac{3}{2}} T}$} & \multirow{2}{*}{P} \\
     \citep{salgiarandom} &  &  &  &  & \\ \hline
     PE & \multirow{2}{*}{$O\rbr{\ln T}$} & \multirow{2}{*}{$\tilde{O}\rbr{T^{\frac{d - \nu}{d}}}$} & \multirow{2}{*}{$O\rbr{\ln^{2 +\alpha} T}$} & \multirow{2}{*}{$O\rbr{\ln T}$} & \multirow{2}{*}{D} \\ 
     \citep{iwazaki2025improvedregretanalysisgaussian} & & & & & \\ \hline
      \textbf{GP-UCB} & \multirow{2}{*}{$O\rbr{1}$} & \multirow{2}{*}{$\tilde{O}\rbr{T^{\frac{d - \nu}{d}}}$} & \multirow{2}{*}{$O\rbr{\ln^{2} T}$} & \multirow{2}{*}{$O\rbr{1}$} & \multirow{2}{*}{D} \\ 
     \textbf{(Our analysis)} & & & & & \\ \hline
     Conjectured Lower Bound & \multirow{2}{*}{N/A} & \multirow{2}{*}{$\Omega\rbr{T^{\frac{d - \nu}{d}}}$} & \multirow{2}{*}{$\Omega(\ln T)$} & \multirow{2}{*}{$\Omega(1)$} & \multirow{2}{*}{N/A} \\
     \citep{vakili2022open} & & & & & \\ \hline
     Lower Bound & \multirow{2}{*}{N/A} & \multirow{2}{*}{$\Omega\rbr{T^{\frac{d - \nu}{d}}}$} & \multirow{2}{*}{$\Omega(1)$} & \multirow{2}{*}{$\Omega(1)$} & \multirow{2}{*}{N/A} \\
     \citep{liregret} & & & & & \\
    \end{tabular}
    \label{tab:nl_cr_compare}
\end{table}

\begin{table}[tb]
    \centering
    \caption{Comparison between existing noiseless algorithms' guarantees for simple regret and our result (adapted from \citep{iwazaki2025improvedregretanalysisgaussian}). 
    In the regrets of GP-UCB+, EXPLOIT+, MVR, and GP-UCB, $\alpha > 0$ and $C > 0$ are an arbitrary fixed constant and a positive constant, respectively.}
    \begin{tabular}{c|c|c|c}
    \multirow{2}{*}{Algorithm} & \multirow{2}{*}{Regret (SE)} & \multirow{2}{*}{Regret (Mat\'ern)} & \multirow{2}{*}{Type} \\
     &  & & \\ \hline \hline
    GP-EI & \multirow{2}{*}{N/A} & \multirow{2}{*}{$\tilde{O}\rbr{T^{-\frac{\min\{1, \nu\}}{d}}}$} & \multirow{2}{*}{D} \\ 
    \citep{bull2011convergence} &  &  & \\ \hline
    GP-EI with $\epsilon$-Greedy & \multirow{2}{*}{N/A} & \multirow{2}{*}{$\tilde{O}\rbr{T^{-\frac{\nu}{d}}}$} & \multirow{2}{*}{P} \\ 
    \citep{bull2011convergence} &  &  & \\ \hline
    GP-UCB & \multirow{3}{*}{$O\rbr{\sqrt{\frac{\ln^{d} T}{T}}}$} & \multirow{3}{*}{$\tilde{O}\rbr{T^{-\frac{\nu}{2\nu + d}}}$} & \multirow{3}{*}{D} \\
    \citep{lyu2019efficient} &  &  & \\
    \citep{kim2024bayesian} &  &  & \\ \hline
    Kernel-AMM-UCB & \multirow{2}{*}{$O\rbr{\frac{\ln^{d+1} T}{T}}$} & \multirow{2}{*}{$\tilde{O}\Bigl(T^{-\frac{\nu d + 2\nu^2}{2\nu^2 + 2\nu d + d^2}}\Bigr)$} & \multirow{2}{*}{D} \\
    \citep{flynn2024tighter} &  &  & \\ \hline
    GP-UCB+, EXPLOIT+ & \multirow{2}{*}{$O\rbr{\exp\rbr{-CT^{\frac{1}{d}-\alpha}}}$} & \multirow{2}{*}{$O\rbr{T^{-\frac{\nu}{d}+\alpha}}$} & \multirow{2}{*}{P} \\ 
    \citep{kim2024bayesian} & & & \\ \hline
    MVR & \multirow{2}{*}{$O\rbr{\exp\rbr{-\frac{1}{2}T^{\frac{1}{d+1}}\ln^{-\alpha} T}} $} & \multirow{2}{*}{$\tilde{O}\rbr{T^{-\frac{\nu}{d}}}$} & \multirow{2}{*}{D}  \\
    \citep{iwazaki2025improvedregretanalysisgaussian} & & &  \\ \hline
    \textbf{GP-UCB} & \multirow{2}{*}{$O\rbr{\sqrt{T}\exp\rbr{-\frac{1}{2} C T^{\frac{1}{d+1}}}} $} & \multirow{2}{*}{$\tilde{O}\rbr{T^{-\frac{\nu}{d}}}$} & \multirow{2}{*}{D}  \\
    \textbf{(Our analysis)} & & & \\ \hline
    Lower Bound & \multirow{2}{*}{N/A} & \multirow{2}{*}{$\Omega\rbr{T^{-\frac{\nu}{d}}}$} & \multirow{2}{*}{N/A} \\
    \citep{bull2011convergence} & & & \\
    \end{tabular}
    \label{tab:nl_sr_compare}
\end{table}

\section{Preliminaries}
\label{sec:prelim}
\paragraph{Noise-free GP bandit problem.} 
We study the GP bandit problem under noise-free observations. We write $\N_+ = \{1,2,\ldots\}$ and $[T] = \{1,\ldots,T\}$ for $T \in \N_+$.
Let $\mX \subset \R^d$ be a compact input domain, and consider a black-box objective function $f:\mX \to \R$ that can only be evaluated point-wise.
At each step $t \in \N_+$, the learner selects a query point $\bx_t \in \mX$ and observes its function value $f(\bx_t)$.
After $T$ steps, the performance of the learner is measured using either the cumulative regret $R_T$ or the simple regret $r_T$, which are respectively defined as follows:
\begin{align}
    R_T &= \sum_{t=1}^{T} [f(\bx^\ast) - f(\bx_t)], \\
    r_T &= f(\bx^\ast) - f(\hat{\bx}_T).
\end{align}
Here, $\bx^\ast \in \mathrm{arg\,max}_{\bx \in \mX} f(\bx)$ is a global maximizer, and $\hat{\bx}_T$ denotes the estimated maximizer returned by the algorithm at the end of the final step $T$.

\paragraph{Gaussian process model.} 
To construct the GP-bandit algorithm, the GP model~\citep{Rasmussen2005-Gaussian} plays a central role in balancing the trade-off between exploration and exploitation. 
First, we adopt a Gaussian process prior with zero mean and covariance (kernel) function $k:\mX \times \mX \to \R$. Then, given a sequence of queried points $\bX_t = (\bx_1,\dots,\bx_t)$ and their corresponding evaluations $\bm{f}(\bX_t) = (f(\bx_1),\dots,f(\bx_t))$, the posterior mean $\mu(\bx;\bX_t)$ and variance $\sigma^2(\bx;\bX_t)$ of $f(\bx)$ at a new point $\bx \in \mX$ are
\begin{align}
    \mu(\bx;\bX_t) &= \bk(\bx, \mE(\bX_t))^\top \bK(\mE(\bX_t), \mE(\bX_t))^{-1} \bm{f}(\mE(\bX_t)), \\
    \sigma^2(\bx;\bX_t) &= k(\bx,\bx) - \bk(\bx, \mE(\bX_t))^\top \bK(\mE(\bX_t), \mE(\bX_t))^{-1} \bk(\bx, \mE(\bX_t)),
\end{align}
where $\bk(\bx,\mE(\bX_t)) = [k(\tilde{\bx}, \bx)]_{\tilde{\bx}\in \mE(\bX_t)}$ is the kernel vector, 
$\bK(\mE(\bX_t), \mE(\bX_t))$ is the Gram matrix, and $\bm{f}(\mE(\bX_t)) = [f(\tilde{\bx})]_{\tilde{\bx} \in \mE(\bX_t)}$.
 In the above definition, $\mE(\bX_t)$ denotes the subset of $\bX_t$ obtained by removing any fully correlated inputs (i.e., with zero posterior variance) with previous ones.
Namely, we define $\mE(\bX_t)$ inductively as $\mE(\bX_t) = \mE(\bX_{t-1}) \cup \{\bx_t\}$ if $\sigma^2(\bx_t; \bX_{t-1}) > 0$; otherwise, $\mE(\bX_t) = \mE(\bX_{t-1})$. 
Here, we define $\mE(\bX_1) = \bX_1$. Note that if there are no duplications in the input sequence $\bx_1, \ldots, \bx_t$, then $\mE(\bX_t) = \bX_t$ holds under commonly used kernel (covariance) functions, such as squared exponential and Mat\'ern kernels (precisely defined in the next paragraph).
Furthermore, for ease of notation, we set $\mu(\bx; \bX) = 0$ and $\sigma^2(\bx; \bX) = k(\bx, \bx)$ for $\bX = \emptyset$.

\paragraph{Kernel function and information gain.}
Regarding the choice of the kernel function, we focus on the squared exponential (SE) kernel
\begin{align}
k_{\mathrm{SE}}(\bx,\tilde{\bx}) &= \exp\Big(-\frac{\|\bx-\tilde{\bx}\|_2^2}{2\ell^2}\Big),
\end{align}
and the Mat\'ern kernel
\begin{align}
k_{\mathrm{Mat\acute{e}rn}}(\bx,\tilde{\bx}) &= 
\frac{2^{1-\nu}}{\Gamma(\nu)} \Big(\frac{\sqrt{2\nu}\,\|\bx-\tilde{\bx}\|_2}{\ell}\Big)^\nu
K_\nu\Big(\frac{\sqrt{2\nu}\,\|\bx-\tilde{\bx}\|_2}{\ell}\Big),
\end{align}
where $\ell>0$ and $\nu>0$ are the lengthscale and smoothness parameter, respectively. Furthermore, $K_\nu$ and $\Gamma$ are the modified Bessel function of the second kind and the gamma function, respectively. 
These two kernels are commonly used and analyzed in GP-bandits~\citep{scarlett2017lower,srinivas10gaussian}. 
The convergence rate of the function estimation in GP regression depends on the choice of kernel, which in turn affects the resulting regret upper bound. Therefore, to capture the problem complexity in a kernel-dependent manner, the following kernel-dependent information-theoretic quantity $\gamma_T(\lambda^2)$ is often employed in the analysis of GP bandits:
\begin{equation}
    \gamma_T(\lambda^2) = \sup_{\bx_1, \ldots, \bx_T \in \mX} \frac{1}{2} \ln \det (\bI_T + \lambda^{-2} \bK(\bX_T, \bX_T)),
\end{equation}
where $\lambda > 0$ and $\bI_T$ are a positive parameter and the $T \times T$ identity matrix, respectively.
The quantity $\gamma_T(\lambda^2)$ is called \emph{maximum information gain} (MIG)~\citep{srinivas10gaussian} since the quantity $\frac{1}{2} \ln \det (\bI_T + \lambda^{-2} \bK(\bX_T, \bX_T))$ represents the mutual information between the underlying function $f$ and training outputs under the noisy-GP model with variance parameter $\lambda^2$. The increasing speed of MIG is analyzed in several commonly used kernels. For example, $\gamma_T(\lambda^2) = O(\ln^{d+1} (T/\lambda^2))$ and 
$\gamma_T(\lambda^2) = \tilde{O}((T/\lambda^2)^{\frac{d}{2\nu+d}})$ under $k = \sek$ and $k = \matk$ with $\nu > 1/2$, respectively~\citep{vakili2021information}\footnote{These orders hold as $T \rightarrow \infty, \lambda \rightarrow 0$.}. 

\paragraph{Regularity assumption.} 
It is hopeless to derive meaningful guarantees without further assumptions about $f$. To obtain a valid estimate of $f$ through the GP model, we assume that $f$ lies in the known reproducing kernel Hilbert space (RKHS)~\citep{aronszajn1950theory} corresponding to the kernel $k$ and has bounded norm. 
The formal description is given below.
\begin{assumption}
    \label{asmp:func}
    The objective function $f$ lies in the reproducing kernel Hilbert space (RKHS) associated with a known positive definite kernel $k: \mX \times \mX \rightarrow \R$. We assume that $k(\bx, \bx) \leq 1$ for all $\bx \in \mX$, and that the RKHS norm $\|f\|_{k}$ satisfies $\|f\|_{k} \leq B < \infty$.
\end{assumption}
This is a standard assumption in the GP-bandit literature~\citep{chowdhury2017kernelized,scarlett2017lower,srinivas10gaussian}, and is leveraged to derive the confidence bound of $f$~\citep{kanagawa2018gaussian,lyu2019efficient,vakili2021optimal}.
Here, the RKHS associated with kernel $k$ is the completion, with respect to the reproducing-kernel inner product, of the linear span $\mathcal{H}_k^{(\mathrm{pre})} \coloneqq \{\sum_{i=1}^n c_i k(\bx^{(i)}, \cdot) \mid n \in \N_+, c_1, \ldots, c_n \in \R, \bx^{(1)}, \ldots, \bx^{(n)} \in \mX\}$~\citep{kanagawa2018gaussian}.
Therefore, intuitively, under Assumption~\ref{asmp:func}, 
we can interpret that the basic properties of the objective function $f$, such as continuity and smoothness, are encoded through the choice of the kernel function $k$.

\paragraph{Gaussian process upper confidence bound.}
GP-UCB~\citep{srinivas10gaussian} is a widely used algorithm for the noisy GP bandit setting. Its noiseless variant was proposed by \citet{lyu2019efficient}, and is outlined in \algoref{alg:gp-ucb}. Their analysis largely follows the framework developed for the noisy case in \citep{srinivas10gaussian}, but differs in the confidence bound, which is strictly tighter than that in the noisy setting.
Although this refinement of the confidence bound leads to superior regret compared with the noisy setting, the resulting regret is still strictly sub-optimal in the noise-free setting. This fact suggests that we require a further fundamental modification of the proof in \citep{srinivas10gaussian}.

\begin{algorithm}[t!]
    \caption{Gaussian process upper confidence bound (GP-UCB) for the noise-free setting.}
    \label{alg:gp-ucb}
    \begin{algorithmic}[1]
        \REQUIRE Compact input domain $\mX \subset \R^d$, Kernel function $k$, and RKHS norm upper bound~$B \in (0, \infty)$.
        \STATE $\bX_0 \leftarrow \emptyset$, $\beta^{1/2} \leftarrow B$.
        \FOR {$t = 1, 2, \ldots$}
            \STATE $\bm{x}_{t} \leftarrow \mathrm{arg~max}_{\bm{x} \in \mathcal{X}} \mu(\bx; \bX_{t-1}) + \beta^{1/2} \sigma(\bx; \bX_{t-1})$.
            \STATE Observe $f(\bx_t)$ and update the posterior mean and variance.
        \ENDFOR
    \end{algorithmic}
\end{algorithm}

\section{Refined Regret Upper Bound for Noise-Free GP-UCB}
\label{sec:reg_ub_gpucb}

The following theorem describes our main results, which show the nearly-optimal regret upper bound for GP-UCB.
\begin{theorem}[Refined cumulative regret upper bound for GP-UCB]
    \label{thm:cr_gp-ucb}
    Fix any compact input domain $\mX \subset \R^d$.
    Suppose $B$, $d$, $\ell$, and $\nu$ are fixed constants.
    Then, when running \algoref{alg:gp-ucb} under \asmpref{asmp:func}, 
    the following two statements hold for any $T \in \N_+$:
    \begin{itemize}
        \item If $k = \sek$, the regret $R_T$ satisfies $R_T = O(1)$.
        \item If $k = \matk$ with $\nu > 1/2$, the regret $R_T$ satisfies
        \begin{equation}
        R_T = 
            \begin{cases}
                \tilde{O}\rbr{T^{\frac{d - \nu}{d}}} ~~&\mathrm{if}~~d > \nu, \\
                O(\ln^2 T) ~~&\mathrm{if}~~d = \nu, \\
                O\rbr{1}  ~~&\mathrm{if}~~d < \nu.
            \end{cases}
        \end{equation}
    \end{itemize}
    The implied constants may depend on $B$, $d$, $\ell$, $\nu$, and the diameter of $\mX$.
\end{theorem}

\begin{theorem}[Refined simple regret upper bound for GP-UCB]
    \label{thm:sr_gp-ucb}
    Fix any compact input domain $\mX \subset \R^d$.
    Suppose $B$, $d$, $\ell$, and $\nu$ are fixed constants.
    Then, when running \algoref{alg:gp-ucb} under \asmpref{asmp:func}, the following two statements hold by setting the estimated maximizer $\hat{\bx}_T$ as 
    $\hat{\bx}_T \in \argmax_{\bx \in \{\bx_1, \ldots, \bx_T\}} f(\bx)$:
    \begin{itemize}
        \item If $k = \sek$, $r_T = O\rbr{\sqrt{T} \exp\rbr{-\frac{1}{2} C T^{\frac{1}{d+1}}}}$.
        \item If $k = \matk$ with $\nu > 1/2$, $r_T = \tilde{O}\rbr{T^{-\frac{\nu}{d}}}$.
    \end{itemize}
    The implied constants depend on $B$, $d$, $\ell$, $\nu$, and the diameter of $\mX$. Furthermore, the constant $C > 0$ depends on $d$, $\ell$, $\nu$, and the diameter of $\mX$\footnote{
    As described in \secref{sec:sketch_pv_ub}, the constant $C$ arises from the implied constants in the upper bound of MIG~\citep{vakili2021information}, which depends on $d$, $\ell$, $\nu$, and the diameter of $\mX$.
    }
    \footnote{Our results (Theorems~\ref{thm:cr_gp-ucb} and \ref{thm:sr_gp-ucb}) heavily rely on the upper bound of the MIG provided in \citep{vakili2021information}, which relies on the uniform boundedness assumption of the eigenfunctions of the kernel.
    The validity of the uniform boundedness assumption is doubted by \citet{janz2022sequential} under a general compact input domain $\mX$. Although our results are based on the upper bound of MIG in \citep{vakili2021information}, our proof strategy is also applicable for deriving nearly-optimal regrets based on the recent analysis of MIG~\citep{iwazaki2025improved} without the uniform boundedness assumption. Specifically, then, the orders of the resulting regrets are the same as those in Theorems~\ref{thm:cr_gp-ucb} and \ref{thm:sr_gp-ucb} for $\sek$ and $\matk$ with $d < \nu$. As for the case under $k = \matk$ with $d \geq \nu$, we can also obtain $R_T =\tilde{O}(T^{(d-\nu)/d})$ and $r_T = \tilde{O}(T^{-\nu/d})$, while they suffer from additional polylogarithmic terms.
    }
    .
\end{theorem}

\paragraph{Proof sketch.}
Our key technical results are the new analysis of the cumulative posterior standard deviation $\sum_{t=1}^T \sigma(\bx_t; \bX_{t-1})$ and its minimum $\min_{t \in [T]} \sigma(\bx_t; \bX_{t-1})$, which play an important role in the theoretical analysis of GP bandits.
Indeed, following the standard analysis of GP-UCB, we have the following upper bounds of regrets by combining the UCB-selection rule with the existing noise-free confidence bound (e.g., Lemma~11 in \citep{lyu2019efficient} or Proposition~1 in \citep{vakili2021optimal}):
\begin{align}
    \label{eq:Rt_ub}
    R_T &= \sum_{t=1}^T [f(\bx^{\ast}) - f(\bx_t)] \leq 2B \sum_{t=1}^T \sigma(\bx_t; \bX_{t-1}), \\
    \label{eq:rt_ub}
    r_T &= \min_{t \in [T]} \{f(\bx^{\ast}) - f(\bx_t)\} \leq 2B \min_{t \in [T]} \sigma(\bx_t; \bX_{t-1}).
\end{align}

From the above inequalities, we observe that the tighter upper bounds of $\sum_{t=1}^T \sigma(\bx_t; \bX_{t-1})$ and $\min_{t \in [T]} \sigma(\bx_t; \bX_{t-1})$ directly yield the tighter regret upper bounds of GP-UCB. 
\lemref{lem:pv_ub} below is our main technical contribution, which gives the refined upper bounds of $\sum_{t=1}^T \sigma(\bx_t; \bX_{t-1})$ and $\min_{t \in [T]} \sigma(\bx_t; \bX_{t-1})$. 

\begin{lemma}[Posterior standard deviation upper bound for SE and Mat\'ern kernel]
    \label{lem:pv_ub}
    Fix any compact input domain $\mX \subset \R^d$, and kernel function $k: \mX \times \mX \rightarrow \R$ that satisfies $k(\bx, \bx) \leq 1$ for all $\bx \in \mX$. Then, the following statements hold for any $T \in \N_+$ and any input sequence $\bx_1, \ldots, \bx_T \in \mX$:
    \begin{itemize}
        \item For $k = \sek$, we have
        \begin{align}
            \label{eq:se_min_cum}
            \min_{t \in [T]} \sigma(\bx_t; \bX_{t-1}) = O\rbr{\sqrt{T} \exp\rbr{-\frac{1}{2}C T^{\frac{1}{d+1}}}}~~\mathrm{and}~~
             \sum_{t=1}^T \sigma(\bx_t; \bX_{t-1}) = O(1).
        \end{align}
        \item For $k = \matk$ with $\nu > 1/2$, we have
        \begin{align}
            \min_{t \in [T]} \sigma(\bx_t; \bX_{t-1}) &= O\rbr{T^{-\frac{\nu}{d}} \ln^{\frac{\nu}{d}} T}, \\
            \label{eq:mat_cum_bound}
            \sum_{t=1}^T \sigma(\bx_t; \bX_{t-1}) &= \begin{cases}
            O\rbr{T^{\frac{d-\nu}{d}} \ln^{\frac{\nu}{d}} T} ~~&\mathrm{if}~~d > \nu, \\
            O\rbr{\ln^{2} T} ~~&\mathrm{if}~~d = \nu, \\
            O(1)  ~~&\mathrm{if}~~d < \nu.
            \end{cases}
        \end{align}
    \end{itemize}
    The constant $C > 0$ and the implied constants depend on $d$, $\ell$, $\nu$, and the diameter of $\mX$. 
\end{lemma}


The full proof of \lemref{lem:pv_ub} is given in Appendix~\ref{subsec:proof_pv_ub}. We will also provide the proof sketch in the next section.
Combining the above equations with Eqs.~\eqref{eq:Rt_ub} and \eqref{eq:rt_ub}, we obtain the statements in Theorems~\ref{thm:cr_gp-ucb} and~\ref{thm:sr_gp-ucb}.

\paragraph{Relation to the existing research in noise-free setting.}
\lemref{lem:pv_ub} resolves the open problem raised by the existing noise-free setting literature~\citep{liregret,vakili2022open}.
First, \citet{vakili2022open} conjectured that the quantity $\sum_{t=1}^T \sigma(\bx_t; \bX_{t-1})$ under $k = \matk$ can attain the following upper bound:
\begin{align}
    \label{eq:conjecture}
    \sum_{t=1}^T \sigma(\bx_t; \bX_{t-1}) &= \begin{cases}
    O\rbr{T^{\frac{d-\nu}{d}}} ~~&\mathrm{if}~~d > \nu, \\
    O\rbr{\ln T} ~~&\mathrm{if}~~d = \nu, \\
    O(1)  ~~&\mathrm{if}~~d < \nu.
    \end{cases}
\end{align}
Although the above conjecture is partially validated under a specific non-adaptive algorithm~\citep{liregret,iwazaki2025improvedregretanalysisgaussian,salgiarandom}, the correctness of this conjecture under a general algorithm has been an open problem~\citep{liregret}. \lemref{lem:pv_ub} answers this open problem with Eq.~\eqref{eq:mat_cum_bound}, which matches the conjectured upper bound up to polylogarithmic factors.

\paragraph{Generality of \lemref{lem:pv_ub}.}
    We would like to highlight that \lemref{lem:pv_ub} always holds for any input sequence, in contrast to the existing algorithm-specific upper bounds~\citep{iwazaki2025improvedregretanalysisgaussian,salgiarandom}. Since the existing noisy GP bandits theory often uses the upper bound of $\min_{t \in [T]} \sigma(\bx_t; \bX_{t-1})$ or $\sum_{t=1}^T \sigma(\bx_t; \bX_{t-1})$ from \citep{srinivas10gaussian}, we expect that many existing theoretical results in the noisy setting can be extended to the corresponding noise-free setting by directly replacing the existing noisy upper bounds of \citep{srinivas10gaussian} with \lemref{lem:pv_ub}. Examples include analyses of GP-Thompson sampling (GP-TS)~\citep{chowdhury2017kernelized}, GP-UCB and GP-TS under the Bayesian setting~\citep{Russo2014-learning,srinivas10gaussian}, the contextual setting~\citep{krause2011contextual}, GP-based level-set estimation~\citep{gotovos2013active}, the multi-objective setting~\citep{JMLR:v17:15-047}, and the robust formulation~\citep{bogunovic2018adversarially}.

\paragraph{Extension to other kernel functions.}
\lemref{lem:pv_ub} is limited to the SE and Mat\'ern kernels. 
However, our proof strategy in \secref{sec:sketch_pv_ub} can be applied to any kernel function if we know the joint dependence of $T$ and the noise-variance parameter $\lambda^2$ in MIG. For example, we can apply our proof strategy for the neural tangent kernel (NTK)~\citep{jacot2018neural} by leveraging the existing upper bound of MIG under NTK~\citep{iwazaki2024no,kassraie2022neural,kassraie2022graph,vakili2021uniform}. 
We expect that such an extension will benefit the theoretical guarantees of neural network-based bandit algorithms under a noise-free setting and will be one of the interesting research directions.

\subsection{Proof Sketch of \lemref{lem:pv_ub}}
\label{sec:sketch_pv_ub}
In this subsection, we describe the proof sketch of \lemref{lem:pv_ub}, while we give its full proof in Appendix~\ref{subsec:proof_pv_ub}. 
Below, to prove \lemref{lem:pv_ub}, we consider the more general lemmas, which bridge the MIG to noise-free posterior standard deviations. 

\begin{lemma}[General upper bound for the minimum posterior standard deviation]
    \label{lem:mpv_ub}
    Fix any input domain $\mX$ and any integer $\overline{T} \geq 2$.
    Let $(\lambda_t)_{t \geq \overline{T}}$ be a strictly positive sequence such that $\gamma_t(\lambda_t^2) \leq (t-1)/3$ for all $t \geq \overline{T}$.
    Then, $\min_{t \in [T]} \sigma(\bx_t; \bX_{t-1}) \leq \lambda_T$ holds for any $T \geq \overline{T}$ and any sequence $\bx_1, \ldots, \bx_T \in \mX$. 
\end{lemma}

\begin{lemma}[General upper bound for the cumulative posterior standard deviations]
    \label{lem:cpv_ub}
    Fix any input domain $\mX$, any integer $\overline{T} \geq 2$, and any kernel function $k: \mX \times \mX \rightarrow \R$ that satisfies $k(\bx, \bx) \leq 1$ for all $\bx \in \mX$.
    Let $(\lambda_t)_{t \geq \overline{T}}$ be a strictly positive sequence such that $\gamma_t(\lambda_t^2) \leq (t-1)/3$ for all $t \geq \overline{T}$.
    Then, the following inequality holds for any $T \in \N_+$ and any sequence $\bx_1, \ldots, \bx_T \in \mX$:
    \begin{equation}
        \label{eq:gen_cpv_ub}
        \sum_{t=1}^T \sigma(\bx_t; \bX_{t-1}) \leq \overline{T} - 1 + \sum_{t=\overline{T}}^T \lambda_t.
    \end{equation}
\end{lemma}
These lemmas hold for any kernel $k$ satisfying $k(\bx, \bx) \leq 1$, which includes most standard kernels after normalization.
Roughly speaking, the above lemmas suggest that $\min_{t \in [T]} \sigma(\bx_t; \bX_{t-1}) \lesssim \lambda_T$ and $\sum_{t=1}^T \sigma(\bx_t; \bX_{t-1}) \lesssim \sum_{t=1}^T \lambda_t$ hold provided that the corresponding MIG $\gamma_t(\lambda_t^2)$ 
does not increase super-linearly. Note that the MIG $\gamma_t(\lambda_t^2)$ monotonically increases as $\lambda_t^2$ decreases, which implies that the tightest upper bound is obtained by setting $\lambda_t^2$ as $\gamma_t(\lambda_t^2) = (t-1)/3$. 
By relying on the existing upper bound of MIG~\citep{vakili2021information}, we can confirm that the condition $\forall t \geq \overline{T}, \gamma_t(\lambda_t^2) \leq (t-1)/3$ of the above lemmas holds with $\lambda_t^2 = O(t \exp(-C t^{\frac{1}{d+1}}))$ and $\lambda_t^2 = O(t^{-\frac{2\nu}{d}} (\ln t)^{\frac{2\nu}{d}})$ for $k=\sek $ and $k = \matk$, respectively. Here, the constants $C$ and $\overline{T}$ are determined based on the implied constant of the upper bound of MIG.
See \appref{subsec:proof_pv_ub} for details. 
\lemref{lem:pv_ub} follows from the aforementioned setting of $\lambda_t^2$, and Lemmas~\ref{lem:mpv_ub} and \ref{lem:cpv_ub}.  Below, we give the proofs for Lemmas~\ref{lem:mpv_ub} and \ref{lem:cpv_ub}.

\paragraph{Proof of \lemref{lem:mpv_ub}.}
Instead of directly treating the noise-free posterior standard deviation, we study its upper bound with the posterior standard deviation of some noisy GP model. Here, let $\sigma_{\lambda^2}^2(\bx; \bX_{t-1})$ denote the posterior variance under the noisy GP-model with the strictly positive variance parameter $\lambda^2 > 0$, which is defined as
\begin{equation}
    \sigma_{\lambda^2}^2(\bx; \bX_{t-1}) = k(\bx, \bx) - \bk(\bx, \bX_{t-1})^{\top} [\bK(\bX_{t-1}, \bX_{t-1}) + \lambda^2 \bI_{t-1}]^{-1}\bk(\bx, \bX_{t-1}).
\end{equation}
Since the posterior variance is monotonic for the variance parameter, 
we have $\sigma^2(\bx_t; \bX_{t-1}) \leq \sigma_{\lambda_T^2}^2(\bx_t; \bX_{t-1})$ for all $t \in [T]$.
Next, we obtain the upper bound of $\sigma_{\lambda_T^2}^2(\bx; \bX_{t-1})$ based on the following lemma, which is the main component of the proof of Lemmas~\ref{lem:mpv_ub} and \ref{lem:cpv_ub}.
\begin{lemma}[Elliptical potential count lemma, Lemma~D.9 in \citep{flynn2024tighter} or Lemma 3.3 in \citep{iwazaki2025improvedregretanalysisgaussian}]
    \label{lem:epcl}
    Fix any $T \in \N_+$, any sequence $\bx_1, \ldots, \bx_T \in \mX$, and $\lambda > 0$.
    Define $\mT$ as $\mT = \{t \in [T] \mid \lambda^{-1} \sigma_{\lambda^2}(\bx_{t}; \bX_{t-1}) > 1\}$, where $\bX_{t-1} = (\bx_1, \ldots, \bx_{t-1})$.
    Then, the number of elements of $\mT$ satisfies $|\mT| \leq 3\gamma_T(\lambda^2)$.
\end{lemma}
The above lemma implies that the set $\mT^c \coloneqq \{t \in [T] \mid \sigma_{\lambda_T^2}(\bx_{t}; \bX_{t-1}) \leq \lambda_T\}$ 
satisfies $|\mT^c| = |[T]\setminus \mT| \geq T - 3\gamma_T(\lambda_T^2)$. Therefore, for any $T \geq \overline{T}$, $|\mT^c| \geq 1$ holds from the condition $\gamma_T(\lambda_T^2) \leq (T-1)/3$. 
This implies there exists some $\tilde{t} \in [T]$ such that $\sigma(\bx_{\tilde{t}}; \bX_{\tilde{t}-1}) \leq \sigma_{\lambda_T^2}(\bx_{\tilde{t}}; \bX_{\tilde{t}-1}) \leq \lambda_T$; therefore, $\min_{t\in [T]} \sigma(\bx_t; \bX_{t-1}) \leq \sigma(\bx_{\tilde{t}}; \bX_{\tilde{t}-1}) \leq \lambda_T$ holds for all $T \geq \overline{T}$. \qed

\paragraph{Proof of \lemref{lem:cpv_ub}.}
Overall, the proof strategy of this lemma is to repeatedly apply the 
proof of \lemref{lem:mpv_ub} by leveraging the monotonicity of the posterior variance against training inputs.
First, if $T < \overline{T}$, Eq.~\eqref{eq:gen_cpv_ub} clearly holds from the assumption $\forall \bx \in \mX, k(\bx, \bx) \leq 1$. Hereafter, we focus on $T \geq \overline{T}$.
By following the same argument of \lemref{lem:mpv_ub}, we can confirm that there exists the index $\tilde{t}_T \leq T$ such that $\sigma(\bx_{\tilde{t}_T}; \bX_{\tilde{t}_T-1}) \leq \sigma_{\lambda_T^2}(\bx_{\tilde{t}_T}; \bX_{\tilde{t}_T-1}) \leq \lambda_T$. Here, we define the new sequence $(\bx_t^{(T-1)})_{t \in [T-1]}$ as the sequence obtained by removing $\bx_{\tilde{t}_T}$ from $(\bx_t)_{t \in [T]}$; namely, we set $\bx_t^{(T-1)} = \1\{t < \tilde{t}_T\} \bx_t + \1\{t \geq \tilde{t}_T\} \bx_{t+1}$ for any $t \in [T-1]$. Furthermore, we define $\bX_t^{(T-1)} = (\bx_1^{(T-1)}, \ldots, \bx_t^{(T-1)})$. From this construction of $\bX_t^{(T-1)}$, we can observe the following two facts:
\begin{itemize}
    \item For any $t < \tilde{t}_T$, we have $\sigma(\bx_t; \bX_{t-1}) = \sigma\rbr{\bx_t^{(T-1)}; \bX_{t-1}^{(T-1)}}$, since $\bx_{t}^{(T-1)} = \bx_t$ and $\bX_{t-1}^{(T-1)} = \bX_{t-1}$.
    \item For any $t > \tilde{t}_T$, we have $\sigma(\bx_t; \bX_{t-1}) \leq \sigma\rbr{\bx_{t-1}^{(T-1)}; \bX_{t-2}^{(T-1)}}$, since $\bx_{t-1}^{(T-1)} = \bx_t$ and $\bX_{t-2}^{(T-1)} \subset \bX_{t-1}$ from the definition of $\bX_t^{(T-1)}$.
\end{itemize}
From the above two facts, we have
\begin{equation}
    \sum_{t=1}^T \sigma(\bx_t; \bX_{t-1}) \leq \sum_{t \in [T] \setminus \{\tilde{t}_T\}} \sigma(\bx_t; \bX_{t-1}) + \lambda_T \leq \sum_{t \in [T-1]} \sigma\rbr{\bx_t^{(T-1)}; \bX_{t-1}^{(T-1)}} + \lambda_T.
\end{equation}

If $T-1 \geq \overline{T}$, we observe that there exists the index $\tilde{t}_{T-1} \leq T-1$ such that $\sigma(\bx_{\tilde{t}_{T-1}}^{(T-1)}; \bX_{\tilde{t}_{T-1}-1}^{(T-1)}) \leq \sigma_{\lambda_{T-1}^2}(\bx_{\tilde{t}_{T-1}}^{(T-1)}; \bX_{\tilde{t}_{T-1}-1}^{(T-1)}) \leq \lambda_{T-1}$ by the application of \lemref{lem:epcl} for the new sequence $(\bx_t^{(T-1)})$. Again, by setting $\bx_t^{(T-2)} = \1\{t < \tilde{t}_{T-1}\} \bx_t^{(T-1)} + \1\{t \geq \tilde{t}_{T-1}\} \bx_{t+1}^{(T-1)}$ and $\bX_t^{(T-2)} = (\bx_1^{(T-2)}, \ldots, \bx_t^{(T-2)})$ for any $t \in [T-2]$, we have
\begin{align}
    \sum_{t \in [T-1]} \sigma\rbr{\bx_t^{(T-1)}; \bX_{t-1}^{(T-1)}} & \leq 
    \sum_{t \in [T-1]\setminus \{\tilde{t}_{T-1}\}} \sigma\rbr{\bx_t^{(T-1)}; \bX_{t-1}^{(T-1)}} + \lambda_{T-1} \\ 
    &\leq 
    \sum_{t \in [T-2]} \sigma\rbr{\bx_t^{(T-2)}; \bX_{t-1}^{(T-2)}} + \lambda_{T-1}.
\end{align}
We can repeat the above arguments until we reach $\overline{T}-1$. Then, the resulting upper bound becomes
\begin{equation}
    \sum_{t=1}^T \sigma(\bx_t; \bX_{t-1}) \leq \sum_{t=1}^{\overline{T}-1} \sigma\rbr{\bx_t^{(\overline{T}-1)}; \bX_{t-1}^{(\overline{T}-1)}} + \sum_{t=\overline{T}}^T \lambda_t \leq \overline{T}-1 + \sum_{t=\overline{T}}^T \lambda_t,
\end{equation}
where the last inequality follows from $\sigma\rbr{\bx_t^{(\overline{T}-1)}; \bX_{t-1}^{(\overline{T}-1)}} \leq \sqrt{k(\bx_t^{(\overline{T}-1)}, \bx_t^{(\overline{T}-1)})} \leq 1$.
\qed

\section{Discussion}
\label{sec:discuss}
Below, we discuss the remaining open questions in the noise-free setting.
\begin{itemize}
    \item \textbf{Lower bound for squared exponential kernel.} While our results establish near-optimality for the Mat\'ern kernel, the optimal simple regret rate under the SE kernel remains unknown. Since the existing upper bound $O(\ln^{d+1} T)$ for MIG~\citep{vakili2021information} does not match $\Omega(\ln^{d/2} T)$ lower bound\footnote{This is derived from the best-known cumulative regret upper bound $R_T = \tilde{O}(\sqrt{T \gamma_T(\lambda^2)})$~\citep{li2022gaussian,salgia2021domain,valko2013finite}, and lower bound $R_T = \Omega(\sqrt{T \ln^{d/2} T})$ in the noisy setting~\citep{scarlett2017lower}.}, we conjecture that further room for improvement also exists in the noise-free setting. Specifically, the exponent $d+1$ of the MIG is reflected in the denominator of the exponential factors in our regret bound in \thmref{thm:sr_gp-ucb}. Therefore, we conjecture that $O(\sqrt{T} \exp(-CT^{2/d}))$ regret is the best guarantee for the simple regret in the SE kernel.
    
    \item \textbf{Constant cumulative regret in Bayesian setting.}
    By using \lemref{lem:pv_ub}, we can prove the same cumulative regret bound as in \thmref{thm:cr_gp-ucb} up to a logarithmic factor in noise-free GP-UCB or GP-TS in the Bayesian setting~\citep{srinivas10gaussian,Russo2014-learning}. However, the confidence width parameter in the Bayesian setting must scale as $\beta^{1/2} = O(\sqrt{\ln T})$ to construct a valid confidence bound. This leads to $O(\sqrt{\ln T})$ regret in SE and Mat\'ern kernel with $d < \nu$ under the Bayesian setting, whereas the frequentist counterpart guarantees constant $O(1)$ regret (\thmref{thm:cr_gp-ucb}). 
    This is counterintuitive, since, as shown in existing analyses for noisy settings~\citep{scarlett2018tight,srinivas10gaussian}, Bayesian regret often achieves smaller values than the worst-case regret in the frequentist setting. An interesting direction for future work is to either design a Bayesian algorithm with constant regret or prove an $\Omega(\sqrt{\ln T})$ lower bound in the Bayesian setting. 

    \item \textbf{GP-UCB in simple regret minimization.} 
    Our analysis shows that GP-UCB achieves nearly optimal simple and cumulative regrets under the Mat\'ern kernel. On the other hand, several algorithms have been proposed that focus on minimizing the simple regret. One of the most well-known examples is EI, which greedily minimizes the simple regret under a Bayesian modeling assumption of GP, and has demonstrated good empirical performance in various applications~\citep{brochu2010tutorial,snoek2012practical}. Based on our theoretical results, there exists no algorithm that can achieve strictly better simple regret than that of GP-UCB in the worst-case sense. Nevertheless, as far as we are aware, GP-UCB tends to exhibit inferior empirical performance compared to EI in simple regret minimization. See Appendix~\ref{sec:ei_ucb}. It remains unclear whether this phenomenon arises from constant factors or additional logarithmic terms in the theoretical regret upper bounds, or whether it reflects a more fundamental gap between worst-case analysis and empirical behavior.
\end{itemize}

\section{Conclusion}
This paper shows that GP-UCB achieves nearly-optimal regret by proving a new regret upper bound for noise-free GP bandits. The key theoretical component of our analysis is a tight upper bound on the posterior standard deviations of GP tailored to a noise-free setting (\lemref{lem:pv_ub}). As remarked in \secref{sec:reg_ub_gpucb}, \lemref{lem:pv_ub} can be applicable beyond the analysis of GP-UCB. Specifically, we expect that many existing theoretical results for noisy GP bandit settings can be translated to the noise-free setting by replacing the existing noisy upper bound of the posterior standard deviations with \lemref{lem:pv_ub}.
For this reason, we believe that our result marks an important step toward advancing the theory for noise-free GP bandit algorithms.

\bibliographystyle{plainnat}
\bibliography{main}

\newpage
\appendix

\onecolumn

\section{Proofs for \secref{sec:reg_ub_gpucb}}
\label{sec:proof_reg_ub_gpucb}
\subsection{Proof of Theorems~\ref{thm:cr_gp-ucb} and \ref{thm:sr_gp-ucb}}

We first formally describe the existing noise-free confidence bound.
\begin{lemma}[Deterministic confidence bound for noise-free setting, e.g., Corollary 3.11 in \citep{kanagawa2018gaussian}, Lemma~11 in \citep{lyu2019efficient}, or Proposition~1 in \citep{vakili2021optimal}]
    \label{lem:detrm_cb}
    Suppose Assumption~\ref{asmp:func} holds.
    Then, for any sequence $(\bx_t)_{t \in \N_+}$ on $\mX$, the following statement holds:
    \begin{equation}
        \forall t \in \N_+,~\forall \bx \in \mX,~|f(\bx) - \mu(\bx; \bX_t)| \leq B \sigma(\bx; \bX_t),
    \end{equation}
    where $\bX_t = (\bx_1, \ldots, \bx_t)$.
\end{lemma}

Although the remaining parts of the proofs are well-known results of GP-UCB, we provide the details for completeness.
Based on the above lemma, we show Eqs.~\eqref{eq:Rt_ub} and \eqref{eq:rt_ub}.
Regarding $R_T$, we have
\begin{align}
    R_T &= \sum_{t=1}^T f(\bx^{\ast}) - f(\bx_t) \\
    &\leq \sum_{t=1}^T [\mu(\bx^{\ast}; \bX_{t-1}) + B \sigma(\bx^{\ast}; \bX_{t-1})] - [\mu(\bx_t; \bX_{t-1}) - B \sigma(\bx_t; \bX_{t-1})] \\
    &\leq \sum_{t=1}^T [\mu(\bx_t; \bX_{t-1}) + B \sigma(\bx_t; \bX_{t-1})] - [\mu(\bx_t; \bX_{t-1}) - B \sigma(\bx_t; \bX_{t-1})] \\
    &= 2B \sum_{t=1}^T \sigma(\bx_t; \bX_{t-1}),
\end{align}
where the first inequality follows from \lemref{lem:detrm_cb}, 
and the second inequality follows from the UCB-selection rule for $\bx_t$. Similarly to the case of cumulative regret, we have
\begin{align}
    r_T &= f(\bx^{\ast}) - f(\hat{\bx}_T) \\
    &\leq \min_{t \in [T]} f(\bx^{\ast}) - f(\bx_t)  \\
    &\leq \min_{t \in [T]} [\mu(\bx^{\ast}; \bX_{t-1}) + B \sigma(\bx^{\ast}; \bX_{t-1})] - [\mu(\bx_t; \bX_{t-1}) - B \sigma(\bx_t; \bX_{t-1})]  \\
    &\leq \min_{t \in [T]} [\mu(\bx_t; \bX_{t-1}) + B \sigma(\bx_t; \bX_{t-1})] - [\mu(\bx_t; \bX_{t-1}) - B \sigma(\bx_t; \bX_{t-1})]  \\
    &= 2B \min_{t \in [T]} \sigma(\bx_t; \bX_{t-1}),
\end{align}
where the first inequality follows from the definition 
of $\hat{\bx}_T$. Finally, the desired results are obtained by combining the above inequalities with Eqs.~\eqref{eq:se_min_cum}--\eqref{eq:mat_cum_bound}. \qed

\subsection{Proof of \lemref{lem:pv_ub}}
\label{subsec:proof_pv_ub}

We prove the following \lemref{lem:pv_ub_detail}, which is a detailed version of \lemref{lem:pv_ub} including explicit constants.
\begin{lemma}[Detailed version of posterior standard deviation upper bound for SE and Mat\'ern kernel]
    \label{lem:pv_ub_detail}
    Fix any compact input domain $\mX \subset \R^d$, and kernel function $k: \mX \times \mX \rightarrow \R$ that satisfies $k(\bx, \bx) \leq 1$ for all $\bx \in \mX$. Furthermore, let $\cse$, $\cmat$, $\ulse$, $\ulmat > 0$, and integers $\utse$, $\utmat \geq 2$ be the constants\footnote{The existence of these constants is guaranteed by the upper bound of MIG~\citep{vakili2021information}, which shows $\gamma_T(\lambda^2) = O(\ln^{d+1} (T/\lambda^2))$ and $\gamma_T(\lambda^2) = O((T/\lambda^2)^{\frac{d}{2\nu+d}} \ln^{\frac{2\nu}{2\nu+d}} (T/\lambda^2))$ (as $T \rightarrow \infty, \lambda \rightarrow 0$) under $k = \sek$ and $k = \matk$, respectively. Note that these constants do not depend on $T$, but may depend on $d$, $\ell$, $\nu$, and the diameter of $\mX$.} that satisfy $\forall \lambda \in (0, \ulse], \forall t \geq \utse, \gamma_t(\lambda^2) \leq \cse (\ln (t/\lambda^2))^{d+1}$
    and $\forall \lambda \in (0, \ulmat], \forall t \geq \utmat, \gamma_t(\lambda^2) \leq \cmat (t/\lambda^2)^{\frac{d}{2\nu+d}}(\ln (t/\lambda^2))^{\frac{2\nu}{2\nu+d}}$ for $k = \sek$ and $k = \matk$, respectively.
    Then, the following statements hold for any $T \in \N_+$ and any input sequence $\bx_1, \ldots, \bx_T \in \mX$:
    \begin{itemize}
        \item For $k = \sek$,
        \begin{align}
            \label{eq:se_min}
            \min_{t \in [T]} \sigma(\bx_t; \bX_{t-1}) &\leq 
            \begin{cases}
                1 ~~&\mathrm{if}~~ T < \tse, \\
                \sqrt{T} \exp\rbr{-\frac{1}{2}\cset T^{\frac{1}{d+1}}} ~~&\mathrm{if}~~ T\geq \tse,
            \end{cases} \\
            \label{eq:se_cum}
             \sum_{t=1}^T \sigma(\bx_t; \bX_{t-1}) &\leq \tse + (d+1) \rbr{\frac{\cset}{2}}^{-\frac{3d+3}{2}} \Gamma\rbr{\frac{3d + 3}{2}},
        \end{align}
        where $\cset = (6\cse)^{-\frac{1}{d+1}}$ and $\tse = \max\{\utse, \utlse, \lceil (d+1)^{d+1}/\cset^{d+1} \rceil + 1\}$ with $\utlse = \min\{T \in \N_+ \mid \forall t \geq T, t\exp(-\cset t^{1/(d+1)}) \leq \ulse^2\}$.
        \item For $k = \matk$ with $\nu > 1/2$,
        \begin{align}
            \label{eq:mat_min}
            \min_{t \in [T]} \sigma(\bx_t; \bX_{t-1}) &\leq \begin{cases}
                1 ~~&\mathrm{if}~~ T < \tmat, \\
                \cmatt^{1/2} T^{-\frac{\nu}{d}} (\ln T)^{\frac{\nu}{d}}  ~~&\mathrm{if}~~ T\geq \tmat,
            \end{cases} \\
            \label{eq:mat_cum}
            \sum_{t=1}^T \sigma(\bx_t; \bX_{t-1}) &\leq \begin{cases}
                \tmat  + \cmatt^{1/2} \frac{d}{d-\nu} T^{\frac{d-\nu}{d}} (\ln T)^{\frac{\nu}{d}} ~~&\mathrm{if}~~d > \nu, \\
                \tmat + \cmatt^{1/2} (\ln T)^{2} ~~&\mathrm{if}~~d = \nu, \\
                \tmat + \cmatt^{1/2} \frac{\Gamma(\frac{\nu}{d} + 1)}{\rbr{\frac{\nu}{d}-1}^{\frac{\nu}{d}+1}} ~~&\mathrm{if}~~d < \nu,
            \end{cases}
        \end{align}
        where $\cmatt = \max\cbr{1, \rbr{2 + \frac{2\nu}{d}}^{\frac{2\nu}{d}} (6\cmat)^{1+\frac{2\nu}{d}}}$ and $\tmat = \max\{4, \utmat, \utlmat\}$ with $\utlmat = \min\{T \in \N_+ \mid \forall t \geq T, \cmatt t^{-\frac{2\nu}{d}} (\ln t)^{\frac{2\nu}{d}} \leq \ulmat^2\}$.
    \end{itemize}
\end{lemma}

The upper bound in \lemref{lem:pv_ub_detail} depends on the quantities $\cset$, $\tse$, $\cmatt$, $\tmat> 0$, which are related to the implied constants in the upper bounds of the MIG. However, under the fixed $d$, $\ell$, and $\nu$, $\cset$, $\tse$, $\cmatt$, $\tmat> 0$ are also constants, which implies the conclusions of \lemref{lem:pv_ub}.

\begin{proof}[Proof of \lemref{lem:pv_ub_detail}]
    When $k = \sek$, we set $\lambda_t^2 = t \exp(-\cset t^{\frac{1}{d+1}})$, $\overline{T} = \tse \coloneqq \max\{\utse, \utlse, \lceil (d+1)^{d+1}/\cset^{d+1} \rceil + 1\}$. 
    From the definition of $\lambda_t^2$ and $\tse$, for any $t \geq \tse$, we have
    \begin{align}
        \gamma_t(\lambda_t^2) 
        &\leq \cse \sbr{\ln\rbr{\frac{t}{\lambda_t^2}}}^{d+1} \\
        &= \cse \sbr{\ln \exp\rbr{\cset t^{\frac{1}{d+1}}}}^{d+1} \\
        &= \cse \cset^{d+1} t.
    \end{align}
    Furthermore,
    \begin{align}
        \cse \cset^{d+1} t \leq \frac{t-1}{3} 
        &\Leftrightarrow \cset^{d+1} \leq \frac{t-1}{3\cse t} \\
        &\Leftarrow \cset^{d+1} \leq \frac{1}{6 \cse} \\
        &\Leftrightarrow \cset \leq \rbr{\frac{1}{6 \cse}}^{\frac{1}{d+1}},
    \end{align}
    where the second line follows from the inequality $t - 1 \geq t/2$ for all $t \geq \tse \geq 2$. By noting the definition of $\cset$, we conclude that 
    $\forall t \geq \tse, \gamma_t(\lambda_t^2) \leq \cse \cset^{d+1} t \leq \frac{t-1}{3}$ from the above inequalities, which implies that Lemmas~\ref{lem:mpv_ub} and \ref{lem:cpv_ub} hold with $\lambda_t^2 = t \exp(-\cset t^{\frac{1}{d+1}})$ and $\overline{T} = \tse$. Eq.~\eqref{eq:se_min} directly follows from Lemma~\ref{lem:mpv_ub} using the fact $\sigma(\bx_t; \bX_{t-1}) \leq \sqrt{k(\bx_t, \bx_t)} \leq 1$. As for 
    Eq.~\eqref{eq:se_cum}, if $T < \tse$, the claimed bound follows from $\sum_{t=1}^T \sigma(\bx_t; \bX_{t-1}) \leq T < \tse$. For $T \geq \tse$, Lemma~\ref{lem:cpv_ub} implies
    \begin{align}
        \sum_{t=1}^T \sigma(\bx_t; \bX_{t-1}) 
        &\leq \tse + \sum_{t=\tse}^{T} \lambda_t \\
        &\leq \tse + \int_{\tse-1}^T \sqrt{t} \exp\rbr{-\frac{1}{2}\cset t^{\frac{1}{d+1}}} \text{d}t\\
        &\leq \tse + \int_{1}^T \sqrt{t} \exp\rbr{-\frac{1}{2}\cset t^{\frac{1}{d+1}}} \text{d}t,
    \end{align}
    where the second line follows from the fact that the function $g(t) \coloneqq t\exp(-\cset t^{1/(d+1)})$ is non-increasing for $t \geq \tse - 1$. In fact, we have
    \begin{align}
        g'(t) = \exp\rbr{-\cset t^{\frac{1}{d+1}}} \rbr{1 - \frac{\cset}{d+1} t^{\frac{1}{d+1}}},
    \end{align}
    which implies $g'(t) \leq 0$ for $t \geq \tse - 1 \geq (d+1)^{d+1}/\cset^{d+1}$. To bound the quantity $\int_{1}^T \sqrt{t} \exp\rbr{-\frac{1}{2}\cset t^{\frac{1}{d+1}}} \text{d}t$, we further derive the following upper bound with $C \coloneqq \cset/2 > 0$:
    \begin{align}
        \int_{1}^T \sqrt{t} \exp\rbr{-C t^{\frac{1}{d+1}}} \text{d}t 
        &= \int_{C}^{CT^{1/(d+1)}} \rbr{\frac{u}{C}}^{(d+1)/2} e^{-u} (d+1) \rbr{\frac{u}{C}}^d \frac{1}{C} \text{d}u~~ (\because u = Ct^{1/(d+1)}) \\
        &= (d+1) C^{-(3d+3)/2}\int_{C}^{CT^{1/(d+1)}} u^{(3d+1)/2} e^{-u} \text{d}u \\
        &\leq (d+1) C^{-(3d+3)/2} \int_{0}^{\infty} u^{(3d+1)/2} e^{-u} \text{d}u \\
        &= (d+1) C^{-(3d+3)/2} \Gamma\rbr{\frac{3d+3}{2}}.
    \end{align}

    Next, when $k = \matk$, we set $\lambda_t^2 = \cmatt t^{-\frac{2\nu}{d}} (\ln t)^{\frac{2\nu}{d}}$ and $\overline{T} = \max\{4, \utmat, \utlmat\}$ with $\cmatt = \max\cbr{1, \rbr{2 + \frac{2\nu}{d}}^{\frac{2\nu}{d}} (6\cmat)^{1+\frac{2\nu}{d}}}$. Then, for any $t \geq \tmat$, it holds that
    \begin{align}
        \gamma_t(\lambda_t^2) 
        &\leq \cmat \rbr{\frac{t}{\lambda_t^2}}^{\frac{d}{2\nu+d}} \sbr{\ln\rbr{\frac{t}{\lambda_t^2}}}^{\frac{2\nu}{2\nu+d}} \\
        &= \cmat \cmatt^{-\frac{d}{2\nu+d}} t (\ln t)^{-\frac{2\nu}{2\nu+d}}  \sbr{\ln \rbr{\cmatt^{-1} t^{\frac{d+2\nu}{d}} (\ln t)^{-\frac{2\nu}{d}} }}^{\frac{2\nu}{2\nu+d}} \\
        &= \cmat \cmatt^{-\frac{d}{2\nu+d}} t (\ln t)^{-\frac{2\nu}{2\nu+d}}  \sbr{\ln \rbr{\cmatt^{-1}} + \frac{d + 2\nu}{d} (\ln t) -\frac{2\nu}{d} (\ln \ln t) }^{\frac{2\nu}{2\nu+d}} \\
        &\leq \cmat \cmatt^{-\frac{d}{2\nu+d}} t (\ln t)^{-\frac{2\nu}{2\nu+d}}  \sbr{\frac{2d + 2\nu}{d} (\ln t)}^{\frac{2\nu}{2\nu+d}} \\
        &= \cmat \cmatt^{-\frac{d}{2\nu+d}} t \rbr{\frac{2d + 2\nu}{d}}^{\frac{2\nu}{2\nu+d}}, 
    \end{align}
    where the fourth line follows from $\cmatt \geq 1 \Rightarrow \cmatt \geq 1/t \Leftrightarrow \ln (\cmatt^{-1}) \leq \ln t$ for $t \geq 1$, and $\ln\ln t \geq 0$ for $t \geq \tmat \geq 4 > e$.
    Furthermore, 
    \begin{align}
        \cmat \cmatt^{-\frac{d}{2\nu+d}} t \rbr{\frac{2d + 2\nu}{d}}^{\frac{2\nu}{2\nu+d}} \leq \frac{t-1}{3} 
        &\Leftrightarrow 3 \cmat \frac{t}{t-1} \rbr{\frac{2d + 2\nu}{d}}^{\frac{2\nu}{2\nu+d}} \leq \cmatt^{\frac{d}{2\nu+d}} \\
        &\Leftrightarrow \rbr{\frac{3 \cmat t}{t-1}}^{1 + \frac{2\nu}{d}} \rbr{2 + \frac{2\nu}{d}}^{\frac{2\nu}{d}} \leq \cmatt \\
        &\Leftarrow \rbr{6 \cmat}^{1 + \frac{2\nu}{d}} \rbr{2 + \frac{2\nu}{d}}^{\frac{2\nu}{d}} \leq \cmatt.
    \end{align}
    Combining the above inequalities, we can confirm $\forall t \geq \tmat, \gamma_t(\lambda_t^2) \leq \frac{t-1}{3}$. 
    Therefore, Lemmas~\ref{lem:mpv_ub} and \ref{lem:cpv_ub} hold with $\lambda_t^2 = \cmatt t^{-\frac{2\nu}{d}} (\ln t)^{\frac{2\nu}{d}}$ and $\overline{T} = \tmat$. Here, Eq.~\eqref{eq:mat_min} is a direct consequence of Lemma~\ref{lem:mpv_ub}. As for Eq.~\eqref{eq:mat_cum}, if $T < \tmat$, the claimed bound follows from $\sum_{t=1}^T \sigma(\bx_t; \bX_{t-1}) \leq T < \tmat$. For $T \geq \tmat$, we have
    \begin{align}
        \sum_{t=1}^T \sigma(\bx_t; \bX_{t-1}) 
        &\leq \tmat  + \sum_{t=\tmat}^{T} \lambda_t \\
        &\leq \tmat + \cmatt^{1/2} \int_{\tmat - 1}^{T} t^{-\frac{\nu}{d}} (\ln t)^{\frac{\nu}{d}} \text{d}t \\
        &\leq \tmat + \cmatt^{1/2} \int_{1}^{T} t^{-\frac{\nu}{d}} (\ln t)^{\frac{\nu}{d}} \text{d}t,
    \end{align}
    where the second line follows from the fact that the function $g(t) \coloneqq t^{-\frac{2\nu}{d}} (\ln t)^{\frac{2\nu}{d}}$ is non-increasing for $t \geq \tmat - 1 \geq 3 > e$. Indeed, we have
    \begin{align}
        g'(t) = \frac{2\nu}{d} t^{-\frac{2\nu}{d}-1} (\ln t)^{\frac{2\nu}{d}}\rbr{(\ln t)^{-1} - 1},
    \end{align}
    which implies $g'(t) \leq 0$ for $t \geq e$.
    The desired results are obtained by bounding the quantity $\int_{1}^{T} t^{-\frac{\nu}{d}} (\ln t)^{\frac{\nu}{d}} \text{d}t$ from above.
    When $d > \nu$, we have
    \begin{equation}
        \int_{1}^{T} t^{-\frac{\nu}{d}} (\ln t)^{\frac{\nu}{d}} \text{d}t
        \leq (\ln T)^{\frac{\nu}{d}} \int_1^T t^{-\frac{\nu}{d}} \text{d}t = (\ln T)^{\frac{\nu}{d}} \sbr{\frac{d}{d-\nu} t^{\frac{d-\nu}{d}}}_1^T \leq \frac{d}{d-\nu} T^{\frac{d-\nu}{d}} (\ln T)^{\frac{\nu}{d}}.
    \end{equation}
    When $d = \nu$,
    \begin{equation}
        \int_{1}^{T} t^{-\frac{\nu}{d}} (\ln t)^{\frac{\nu}{d}} \text{d}t \leq (\ln T) \int_1^T t^{-1} \text{d}t = (\ln T)^2.
    \end{equation}
    When $d < \nu$, we have
    \begin{align}
        \int_{1}^T t^{-\frac{\nu}{d}} (\ln t)^{\frac{\nu}{d}} \text{d}t
        &= \int_{0}^{\ln T} e^{-\rbr{\frac{\nu}{d}-1}u} u^{\frac{\nu}{d}} \text{d}u ~~~(\because u = \ln t) \\
        &\leq \int_{0}^{\infty} e^{-\rbr{\frac{\nu}{d}-1}u} u^{\frac{\nu}{d}} \text{d}u \\
        &= \frac{\Gamma(\frac{\nu}{d} + 1)}{\rbr{\frac{\nu}{d}-1}^{\frac{\nu}{d}+1}},
    \end{align}
    where the last line follows from the standard identity for the gamma function: $\int_{0}^{\infty} e^{-\lambda u} u^b \text{d}u = \Gamma(b+1)/\lambda^{b+1}$ for any $\lambda > 0$ and $b > -1$ (e.g., Equation 6.1.1 in \citep{abramowitz1968handbook}). 
\end{proof}

\section{Experimental Details}
\label{sec:exp_detail}
\subsection{Experimental settings for \figref{fig:emp_regret}}
We give the detailed experimental settings used to plot \figref{fig:emp_regret}.
\begin{itemize}
    \item \textbf{Objective function.} We define the true underlying objective function as $f(\cdot) = \sum_{m=1}^{50} c_m k(\bx^{(m)}, \cdot)$, where $c_m \sim \mathrm{Uniform}([-1, 1])$ and $\bx^{(m)} \sim \mathrm{Uniform}([0, 1]^2)$ are independently generated random variables. Note that by the definition of the RKHS, $f \in \mH_k$ with $\|f\|_{k} = \sqrt{\sum_{m=1}^{50} \sum_{\tilde{m}=1}^{50} c_m c_{\tilde{m}} k(\bx^{(m)}, \bx^{(\tilde{m})})}$.
    \item \textbf{Kernel.} We fix the lengthscale parameter $\ell$ as $\ell = 0.25$ in all experiments. We use the same kernel function in the GP model used in the algorithms as the kernel leveraged to generate the objective function.
    \item \textbf{Other parameters.} We define the input domain $\mX$ as the uniformly aligned $50 \times 50$ grid points on $[0, 1]^2$. In all algorithms, we set the confidence width parameter $\beta^{1/2}$ to the exact RKHS norm. Furthermore, we set the initial batch size in PE and REDS to $5$. Finally, we define the common initial point $\bx_1$ for all the algorithms as the uniformly sampled point from $\mX$. 
\end{itemize}
In the above-described setting, we conduct experiments with $3000$ different seeds.

\subsection{Comparison between EI and GP-UCB}
\label{sec:ei_ucb}

\begin{figure}
    \centering
    \includegraphics[width=0.48\linewidth]{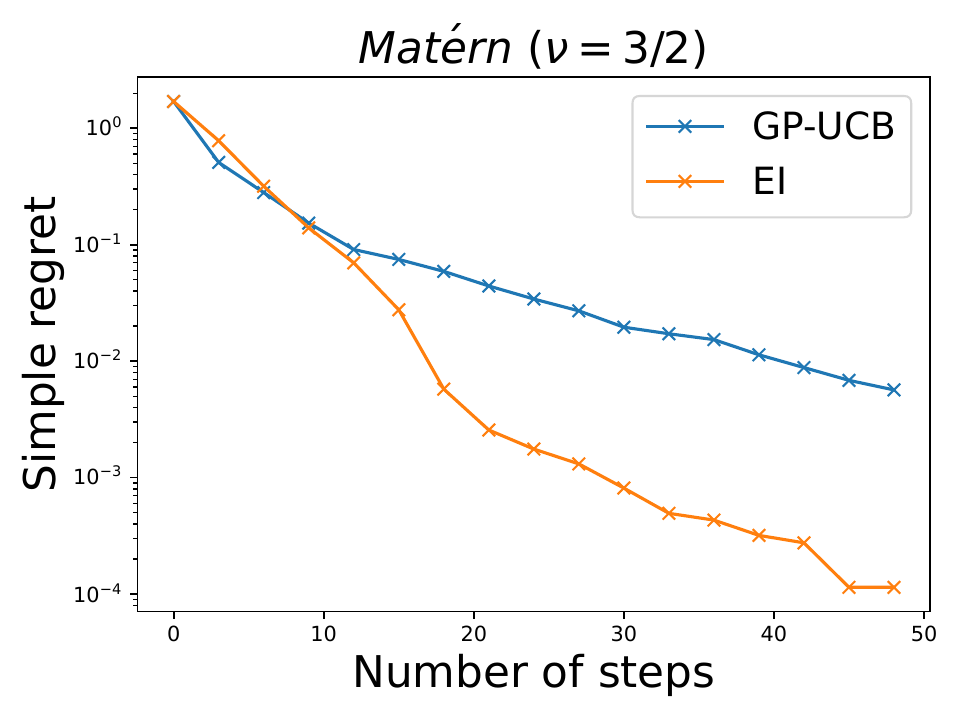}
    \includegraphics[width=0.48\linewidth]{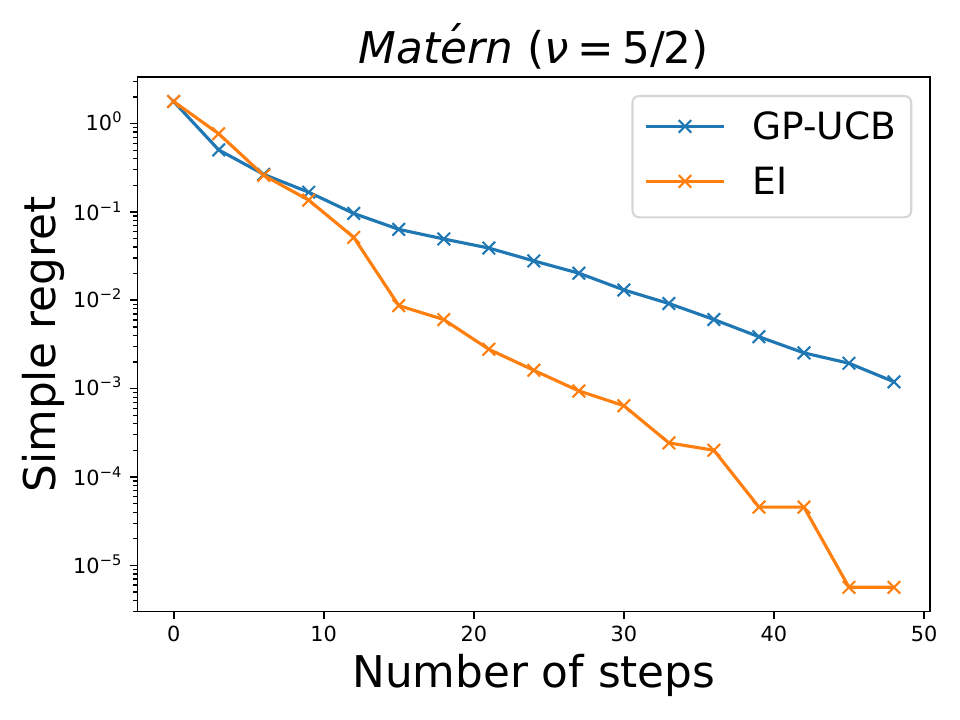}
    \caption{Comparison between GP-UCB and EI in simple regret minimization over $100$ different seeds.
    We conduct experiments with Mat\'ern kernel under $\nu = 3/2$ (left) and $\nu = 5/2$ (right).}
    \label{fig:ei_ucb}
\end{figure}

Under the same setting as the previous subsection, we also compare GP-UCB's empirical performance with that of EI in simple regret minimization. \figref{fig:ei_ucb} shows the results. We can confirm that, although GP-UCB achieves nearly-optimal worst-case regret, the empirical performance of GP-UCB is consistently worse than that of EI.

\end{document}